\documentclass{article}
\usepackage{ijcai21}

\usepackage{times}

\usepackage{soul}
\usepackage{url}
\usepackage[hidelinks]{hyperref}
\usepackage[small]{caption}
\usepackage{graphicx}
\usepackage{booktabs}
\usepackage{amsmath,amssymb,amsfonts}
\usepackage{multirow}

\usepackage{array}
\usepackage{mathrsfs}
\usepackage{subfigure}
\usepackage{epstopdf}
\usepackage{algorithm}
\usepackage{algorithmic}
\usepackage{color}
\usepackage{comment}

\title{MetaMIML:   Meta Multi-Instance Multi-Label Learning}

\author{
Yuanlin Yang$^{1,2}$\and
Guoxian Yu$^{2,3}$\footnote{Contact Author}\and
Jun Wang$^{3}$\and
Lei Liu$^{2}$\and
Carlotta Domeniconi$^4$\and
Maozu Guo$^{5}$\\
\affiliations
$^1$College of Computer and Information Sciences, Southwest University, Chongqing, China\\
$^2$School of Software, Shandong University, Jinan, China\\
$^3$Joint SDU-NTU Centre for Artificial Intelligence Research, Shandong University, Jinan, China\\
$^4$Department of Computer Science, George Mason University, VA, USA\\
$^5$College of Elec. \& Inf. Eng., Beijing Univ. of Civil Eng. and Arch., Beijing, China\\
\emails
ylyang@swu.edu.cn,
\{gxyu, kingjun, l.liu\}@sdu.edu.cn,
carlotta@cs.gmu.edu,
guomaozu@bucea.edu.cn
}

\begin{document}

\maketitle

\begin{abstract}
Multi-Instance Multi-Label learning (MIML) models  complex objects (bags), each of which is associated with a set of interrelated labels and composed with a set of instances. Current MIML solutions still focus on a single-type of objects and assumes an IID distribution of training data. But these objects are linked with objects of other types, 
which also encode the semantics of target objects. In addition, they generally need abundant labeled data for training. To effectively mine interdependent MIML objects of different types, we propose a network embedding and meta learning  based approach (MetaMIML). MetaMIML introduces the context learner with network embedding to capture semantic information of objects of different types, and the task learner to extract the meta knowledge for fast adapting to new tasks. In this way, MetaMIML can naturally deal with MIML objects at data level improving, but also exploit the power of meta-learning at the model enhancing.
Experiments on benchmark datasets demonstrate that  MetaMIML achieves a significantly better performance than state-of-the-art algorithms.
\end{abstract}

\section{Introduction}
In many real-world applications, a complex object of interest has its inherent structure, is represented as a bag of instances and associated with multiple labels simultaneously. Multi-Instance Multi-Label learning (MIML) \cite{Zhou2012MIML} provides a framework for handling such complex bags and instances.
MIML focuses on mining the association between a bag and its instances, labels of bags and label correlations to differentiate the labels of individual instances.
MIML solutions have been extensively applied for many tasks, such as image classification \cite{wu2015deep}, text categorization \cite{forrest2012rank}, gene function prediction \cite{yu2020isoform} and so on. Multi-view MIML (M3L) solutions have been invented to fuse multiple feature views of MIML objects  \cite{nguyen2013M3LDA,yang2018M3DN,xing2019M3Lcmf}, which further model the varying associations between bags and instances across views.

However, these MIML/M3L solutions still model homogeneous bags and neglect these bags linked with objects of other types, which reflect the semantic labels of target bags. As shown in Fig. \ref{fig1} (i), the topic of a research paper in the academic network is not only reflected by its own textual features, but also depends on its authors and published venue. These \emph{multi-types} objects are interdependent, while most MIML methods build on the promise of MIML objects with IID distribution \cite{zhou2008miml,Zhou2012MIML}. To effectively mine the target bags, a natural idea is to encode these bags and their links with other types of objects via a heterogeneous network, and then applies network representation learning to form the composite feature representation of bags or instances. Recently, Multi-types objects Multi-view Multi-instance Multi-label Learning (M4L) is proposed to model interconnected complex objects of different types, and a  joint matrix factorization based solution (M4L-JMF) is introduced \cite{yang2020M4L}. M4L-JMF uses low-rank representation learning on a heterogeneous network composed with multiple types of objects to aggregate the information of other types of objects toward the target bags and to complete the bag/instance-label associations. M4L-JMF proves the necessity and power of fusing multi-types linked MIML objects.

However, existing MIML algorithms and even the more advanced M3L and M4L approaches often require a large amount of training data to achieve good results on such complex dataset, but sufficient labeled bags/instances are hard and even infeasible to collect in practice. Furthermore, they suffer a vulnerable performance when dealing with new tasks, where the training data are even fewer but a good generalization is expected.

\begin{figure*}[h!tbp]
\centering
\includegraphics[width=18cm]{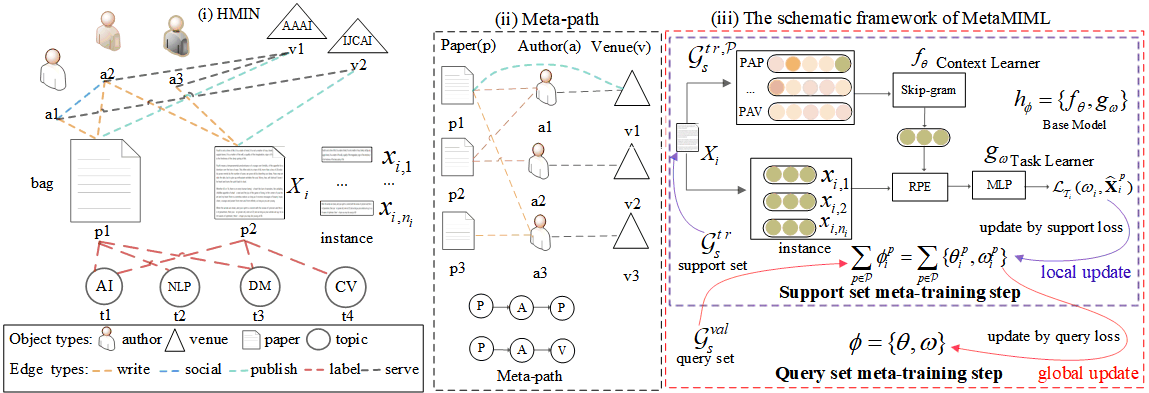}
\caption{(i) An exemplar Heterogeneous Multi-Instance Network (HMIN) composed with papers, authors and published venues, the paper is the complex object further made instances of abstract, references and paragraphs. These papers can be associated with different but related topics (i.e. AI, NLP, CV and DM). (ii) The Meta-path, the meta-path 'A-P-A' indicates two authors have  co-authored  relation.  (iii) The schematic framework of MetaMIML, which first uses the context learner with different meta-paths to learn context embeddings of target objects/instances and to generate multiple cross-tasks (meta-training support samples) from these embeddings. At the same time, it uses task learner to extract meta-knowledge from these tasks and optimizes the embeddings in an iterative manner. The meta-testing phase utilizes the extracted meta-knowledge, embeddings and few meta-test support samples to quickly adopt to new task.}
\label{fig1}
\end{figure*}

Meta learning can extract meta knowledge through learning from multi-related tasks, and then quickly adapt to new tasks with the aid of meta knowledge and few support training samples \cite{vanschoren2019meta}. Although it has achieved good results in various tasks \cite{finn2017model,sung2018learning,snell2017prototypical}, it is still a non-trivial problem to apply meta learning on multi-types MIML objects encoded by heterogeneous network. {\textbf{Challenge 1}: Existing meta learning algorithms are mainly designed for single-type of objects, there is no principle way to apply meta-learning on multi-types MIML objects and to quickly adapt for a new task with just a few training samples.
\textbf{Challenge 2}: How to learn different context representations of bags from heterogeneous multi-instance network for building meta-training tasks and extracting meta knowledge extraction? 
\textbf{Challenge 3}: How to leverage the extracted meta-knowledge and fuse the structure and attribute information of linked objects to predict the labels of target bags/instances.
}

To address the above challenges, we propose an approach called Meta Multi-Instance Multi-Label learning (\textbf{MetaMIML}). MetaMIML firstly constructs a heterogeneous multi-instance network (HMIN) with multiple types of bags, and then trains a context learner and a task learner. Context learner works in the bag contextual semantic space to generate different semantic representations of bags (tasks) from HMIN, while task learner operates in the task space to achieve meta knowledge acquisition from different tasks (for Challenges 1 and 2). MetaMIML then leverage random  embedding and meta knowledge to obtain effective instance/bag embedding representations, and utilizes the task learner to predict the labels of bags/instances (for Challenge 3).


The main contributions of our work are summarized as:
\begin{itemize}
\item We focus on how to integrate meta learning with MIML to quickly adapt to new task with a few labeled MIML objects, which is a practical but largely unexplored topic of meta learning and of MIML. This studied problem has its distinctive challenges and application values.


\item We propose the MetaMIML with a context learner for fusing multiple objects of HMIN and generating multiple context tasks, and with a task learner for acquiring knowledge from multiple context tasks.  MetaMIML not only performs well in  the meta learning setting, but also in the typical MIML settings, by  multi-types objects fusion and meta knowledge.


\item Experimental results on benchmark datasets under different scenarios show that MetaMIML outperforms both the representative MIML algorithms (MIMLfast \cite{huang2019fast}, MIMLSVM \cite{Zhou2012MIML} and MIMLNN \cite{zhou2008miml}), and  matrix factorization based data fusion solutions (MFDF \cite{Zitnik2015DFMF} and SelDFMF \cite{wang2019SelDFMF}) and network embedding-based methods (M4L-JMF \cite{yang2020M4L}, Metapath2vec\cite{dong2017metapath2vec} and HANE \cite{wang2019heterogeneous}).
\end{itemize}


\section{Related work}
\label{sec:relwork}
Our work has close connections with  MIML and Meta Learning. Diverse MIML methods have been proposed in the past decades \cite{Zhou2012MIML}. To name a few, {
MIMLBoost \cite{Zhou2012MIML}  degenerates the MIML problem into multiple multi-instance single-label tasks; and MIMLSVM \cite{Zhou2012MIML} uses a bag transformation strategy to convert bags into single instances and then degenerates MIML into a single-instance multi-label learning task. MIMLNN \cite{zhou2008miml} utilizes  two-layer neural network structure to replace the multi-label SVM used in MIMLSVM.}
Fast Multi-instance Multi-label learning  (MIMLfast) \cite{huang2019fast}  naturally detect key instance for each label by exploiting label relations in a shared space and discovering sub-concepts for complicated labels. Multi-instance multi-label
learning with spatio-temporal pre-trimming (PreTrimNet) \cite{zhang2020multi} learns from the given video-level labels to construct action recognition model, discover the underlying relevance between input patterns (instances) and semantic labels. {
Multi-modal Multi-instance Multi-label LDA (M3LDA) \cite{nguyen2013M3LDA} learns a visual-label part from the visual features and a text-label part from the textual features, and forces these two parts having consistent labels.
Weakly-supervised M3L (WSM3L) \cite{xing2020WSM3L} concerns with unpaired multi-view MIML objects with missing labels by multi-modal dictionary learning.} However, these MIML methods \emph{can only} consider single-type of bags, while in practice, these bags are also linked with objects of other types, which also reflect the semantic labels of target bags.

Heterogeneous network provides a natural way to represent linked objects of multi-types, as such network embedding-based solutions have been proposed  to mine rich semantics between objects by matrix factorization \cite{yu2020attributed,yang2020M4L} or by graph neural networks \cite{dong2017metapath2vec,yang2020HNESurvey}. These solutions aim at mapping the topological proximity of network nodes into continuous low dimensional vector representations for follow-up tasks, such as node classification/clustering, and link prediction.
Although these network embedding based solutions can fuse objects of different types and reduce information bottleneck of target objects at the data level, they ignore that a complex object can be composed of multiple instances, whose feature representations also encode the semantics of target objects.
For example, different salient regions (instances) of a web image often carry different semantics, which give the global semantic of this image.

Meta learning (also called learn to learning) \cite{vanschoren2019meta} aims to use the general knowledge previously learnt from multiple meta-training tasks and a small amount of training data for quickly adapting to new tasks. 
Diverse meta-learning techniques have been invented, and they can be categorized into three types: (i) metric-based methods learn a metric or distance function over tasks \cite{snell2017prototypical,sung2018learning},; (ii) model-based methods \cite{santoro2016meta,munkhdalai2017meta} aim to design an architecture or training process for rapid generalization across tasks; and (iii) optimization-based methods directly adjust the optimization algorithm to enable quick adaptation with just a few examples \cite{finn2017model,hoyeop2019melu}.
These meta learning algorithms still focus on single-type of objects. On the other hand, existing MIML methods performs the optimization of parameter from scratch for every task, and the pre-specified scratch can drastically affect the performance. In contrast, MetaMIML considers multi-types of objects, and it  extracts cross-task knowledge (or meta-knowledge) by a task learner, which help specifying and adaptive optimizing the parameter for target task using just several training samples.



\section{Method}
\label{sec:method}

\subsection{Problem statement and overview}
Let $\mathcal{G} = \{\mathcal{V},\mathcal{E}\}$ be a Heterogeneous Multiple Instance Network (HMIN). $\mathcal{V}$ includes a set of nodes of different types, $\mathcal{E}$ stores links between nodes in $\mathcal{V}$.  HMIN contains at least one type of bags (i.e. papers in Fig. \ref{fig1}(i)),  which  are  further composed with a variable number of instances, $\mathbf{X}_i =\{\mathbf{x}_{i,1},\mathbf{x}_{i,2}, \cdots, \mathbf{x}_{i,n_i}\}$,  where $\mathbf{x}_{i,j} \in \mathbb{R}^d$ is the $j$-instance and $n_i$ is the number of instances affiliated with the $i$-th bag, and $d$ is the dimension  of instance.  In contrast, a typical Heterogeneous Information Network (HIN) only includes objects of different types, or views a multi-instance bag as a plain node, without considering its composition of instances. 
As such, HMIN encodes richer relations among complex objects, instances and labels; it enables a more finer-grained analysis of real-world objects.

MetaMIML aims to explore the fusion of linked objects of different types to enhance the bag/instance representation, and to boost the prediction of labels of target bags/instances. In addition, it has to just leverage a few labeled bags/instances and meta-knowledge to induce an accurate model with good generalization. The optimization-based meta-learning \cite{finn2017model,hoyeop2019melu} optimizes globally shared parameters over tasks, which can rapidly adapt to a new task with just one or few gradient steps. Inspired by this advantage, we build MetaMIML on a set of source tasks as $\mathcal{D}_{s} = \{(\mathcal{G}_{s}^{tr},\mathcal{G}_{s}^{val})^{(i)}\}_{i=1}^{N}$, and a set of target tasks as $\mathcal{D}_{t} = \{(\mathcal{G}_{t}^{tr},\mathcal{G}_{t}^{val})^{(i)}\}_{i=1}^{M}$. Suppose $\mathcal{V}_b \in \mathcal{V}$ denote the target type of bags  and $N+M=n$, $n=|\mathcal{V}_b|$ is the number of bags. More details of data split are explained in Section \ref{sec:isoforms}.

In our MIML setting, a task $ \mathcal{T}^s_i = (\mathcal{G}_{s}^{tr},  \mathcal{G}_{s}^{val})$ involves with one bag, consists of a training  and a validation data. Note that the source training and validation datasets are respectively termed as \emph{support} and \emph{query} sets. We aim to learn meta-knowledge from a set of meta-training tasks $\mathcal{D}_s$, and quickly adapt to new tasks (as meta-test tasks $ \mathcal{D}_t$) using the meta knowledge.  In the meta-training stage:  for each task $ \mathcal{T}^s_i$, its  training  and validation data sampled from the set of target objects (e.g. topics/labels in Fig. \ref{fig1}. The task learner adjusts the global task-wise prior parameters $\phi$ with the loss on the training data $\mathcal{G}_{s}^{tr}$. Next, it calculates the loss with $\phi$  on the validation data $\mathcal{G}_{s}^{val}$, backward propagates the loss to update  $\phi$ as follows:
\begin{equation}
\underset{\phi}{min} \underset{\mathcal{T}^s_i\in \mathcal{D}_{s}}{\mathbb{E}} \mathcal{L}(\mathcal{G}_{s}^{val},\phi -\gamma{\nabla}_{\phi}\mathcal{L}(\mathcal{G}_{s}^{tr},\phi))
\label{eq0-1}
\end{equation}
where  $\mathcal{L}(\mathcal{G}_{s}^{tr},\phi)$ measures the performance of an MIML model trained by $\phi$ on $\mathcal{G}_{s}^{tr}$. $\mathcal{L}$ is the loss function and $\gamma$ is the learning rate. $\phi -\gamma{\nabla}_{\phi}\mathcal{L}(\mathcal{G}_{s}^{tr},\phi)$ is the task parameter adapted for $ \mathcal{T}^s_i $ by one gradient step from $\phi$ ({local update in Fig.  \ref{fig1}(ii)}). {In the meta-test stage, a target task  $\mathcal{T}^t_i \in \mathcal{D}_t$ also contains a small number of training data $\mathcal{G}_{t}^{tr}$, while the validation set  contains $\mathcal{G}_{t}^{val}$.} Conceptually, the base model initialized by meta-knowledge $\phi$ on each unseen target task $\mathcal{T}^t_i$ is fast adapted as follows:
\begin{equation}
\underset{\omega }{min}  \mathcal{L}(\mathcal{G}_{t}^{tr},\omega|\phi)
\label{eq0-2}
\end{equation}
We can evaluate the performance of a meta-learner by $\omega$ on $\mathcal{G}_{t}^{val}$ of each target task. 



\subsection{Context Learner}

Our base model $h_{\phi} = \{f_{\theta} , g_{\omega}\}$ contains two components: \textbf{context learner} $f_{\theta}$ and \textbf{task learner} $g_{\omega}$.  $f_{\theta}$ aims to learn the semantic context information, and $g_{\omega}$ aims to predict the labels of bags/instances. {${\phi} = \{\theta,\omega \}$ denotes the global prior}.

To explore the network structure information of multi-types linked objects and to obtain the context representation of bags, we use the meta-path sampling on the bag-level to capture HMIN.
As a form of higher-order graph structure, meta-path has been widely used to explore the semantics in a HIN \cite{dong2017metapath2vec,yang2020HNESurvey}. {For example the meta-path `A-P-A'/`A-C-A' indicates two authors have co-authored/co-attended a paper/conference.} Given a HMIN with $|\mathcal{V}| = m$ object types and $|\mathcal{E}|$ relation types, a meta-path of length $l$ is defined as a composite relation $p={v}_1 \stackrel{r_1}{\rightarrow} {v}_2 \stackrel{{r_2}}{\rightarrow} \cdots \stackrel{{r}_l}{\rightarrow}{v}_{l+1}$, where $v_i \in \mathcal{V}$ and $r_j \in \mathcal{E}$.

Towards effective meta-learning on HMIN, it is important to incorporate multi-aspect semantic contexts with tasks via different meta-paths. 
Given a task $\mathcal{T}_i^s = (\mathcal{G}_{s}^{tr},\mathcal{G}_{s}^{val})$,
$\mathcal{G}_s^{tr}$ gathers \emph{direct} and \emph{indirect} information of the bag object as:
\begin{equation}
\mathcal{G}_{s}^{tr} = (\mathcal{G}_{i}^{tr,dir} , \mathcal{G}_{i}^{tr,ind}) = (\mathcal{G}_{i}^{tr,dir} , \mathcal{G}_{i}^{tr,\mathcal{P}})
\label{eq1}
\end{equation}
 where $\mathcal{G}_{s,i}^{tr,dir}$ is a set of objects that \emph{directly} connect with objects of type $\mathcal{V}_b$ (bag object),  and $\mathcal{G}_{s,i}^{tr,ind}=  \mathcal{G}_{s,i}^{tr,\mathcal{P}}$ is a set of objects that \emph{indirectly} connect with target objects by meta-paths, $\mathcal{G}_{s,i}^{tr,dir}$ is like a transfer  station for $\mathcal{G}_{i}^{tr,ind}$ to connect different objects.  Since each task $\mathcal{T}^s_i$  may interact with multiple target objects, we build multiple meta-paths based semantic contexts for task $\mathcal{T}^s_i$ as follows:
\begin{equation}
 \mathcal{G}_{i}^{tr,\mathcal{P}} = \{ \mathcal{G}_{i}^{p_1},  \mathcal{G}_{i}^{p_2},\cdots,  \mathcal{G}_{i}^{p_n}\}
\label{eq3}
\end{equation}
where $\mathcal{P} = \{p_1, p_2, \cdots, p_n\}$,  $\mathcal{G}_{i}^{p}$ encodes semantic contexts induced by meta-path $p$. Since we want to obtain the context representation of bags, each meta-path starts with target bags in $\mathcal{V}_b$.
 Therefore, $\mathcal{G}_{s}^{tr,\mathcal{P}}$ contains diverse contextual information from target bags to other objects by walking  along different meta-paths.
Similarly, we can build the query set $\mathcal{G}_{s}^{val} = (\mathcal{G}_{i}^{val,ind} , \mathcal{G}_{i}^{val,\mathcal{P}})$. The support and query set in a task $\mathcal{T}_i^s$ are mutually exclusive, namely $\mathcal{G}_{i}^{tr,dir} \bigcap \mathcal{G}_{i}^{val,dir} = \emptyset$.

In context leaner, the bag information are aggregated from its contexts. Inspired by Word2Vec \cite{mikolov2013efficient}, we use the Skip-Gram model to get the  bag object embedding $\mathbf{X}_{i}$ induced by meta-path $p$ as follows:
\begin{equation}
\mathbf{X}_{i}^p = f_{\theta}(\mathcal{G}_{i}^{p})=\sigma(SG(\mathcal{G}_{i}^{p},\theta))
\label{eq4}
\end{equation}
where $\mathbf{X}_{i}^p \in \mathbb{R}^{d_l}$ and $p \in \mathcal{P}$ . $\sigma$ is the activation function (such as LeaklyReLU). $f_{\theta}$ is the embedding function  and $SG(\cdot)$ is the Skip-Gram model parameterized by $\theta =\{\mathbf{W}\in\mathbb{R}^{l\times d_l}$ (also called look up table), $\mathbf{b}\in\mathbb{R}^{d_l}\}$, $l$ is the length of meta-path $p$ and $d_l$ is the embedding dimension. {To obtain the aggregated representation of instances}, we merge the context representation of bags with instance features through random projection embedding as follows:
\begin{equation}
\hat{\mathbf{x}}_{i,j}^p = (\mathbf{x}_{i,j}\oplus\mathbf{X}_{i}^p)^T \cdot\mathbf{E}
\label{eq5}
\end{equation}
where $\{\mathbf{x}_{i,j}\}_{j=1}^{n_i}$ is the set of instances affiliated with the $i$-th bag, $\oplus$ denotes the concatenation operation,   $\hat{\mathbf{x}}_{i,j}^p \in\mathbb{R}^k$ is the instance representation for the $j$-th instance of this bag, and $k$ is the dimension  of final embedding. $\mathbf{E}\in \mathbb{R}^{u\times k}$ is the sparse random projection matrix and $u=d_l+d$. In this way, we can obtain  a bag's context representation $\hat{\mathbf{X}}_{i}^p= \{\hat{\mathbf{x}}_{i,1}^p,\hat{\mathbf{x}}_{i,2}^p,\cdots,\hat{\mathbf{x}}_{i,n_i}^p  \}$ w.r.t. meta-path $p$ .

Random projection is a simple but powerful technique of dimensionality reduction, which can preserve the pairwise distance between data points. The sparse random projection we use is a variant of random projection, it can achieve a significant speedup with little loss in accuracy. Formally, each entry of $\mathbf{E}\in\mathbb{R}^{u\times k}$ is IID drawn from:
\begin{equation}
e_{ij}= \left\{\begin{array}{c}
    1,   \quad          \text{with prob.  $\frac {1}{s}$}\\
    0, \quad              \text{with prob. $1-\frac {1}{s}$}\\
   -1 ,  \quad          \text{with prob.  $\frac {1}{2s}$}\\
  \end{array}
  \right.
  \label{eq6}
\end{equation}
where $s = \sqrt{n}$ or $s = \frac{n}{logn}$, $n$ is the number of bags. {Eq. \eqref{eq6} can achieve a $\sqrt{n}$-fold speedup, and only  $1/\sqrt{n}$ entries need to be processed when $s = \sqrt{n}$ \cite{li2006very}.}


\subsection{Task Learner}
Through the context learner introduced above, we have obtained an instance representation that combines semantic contexts and instance features. Because different semantic context representations have different degrees of importance for different objects, the change in semantics also changes the representation of instances. Similarly, new tasks have different preferences for semantic representations. Given that, we firstly use the \textbf{task leaner} to predict the labels of bags/instances, then the support loss  locally updates parameters via gradient descent and back-propagation.

In the meta-path $p$ induced semantic space, we can calculate the loss on the support set $\mathcal{G}_s^{tr}$ and task $\mathcal{T}_i^s$ as:
\begin{equation}
 \begin{split}
\mathcal{L}_{\mathcal{T}_i}(\mathbf{\omega},\hat{\mathbf{X}}^p_i, \mathcal{G}_{i}^{tr,dir}) &=  ||\mathbf{y}-g_{\mathbf{\omega}}(\hat{\mathbf{X}}^p_i)||^2 \\
&  +||g_{\mathbf{\omega}}(\hat{\mathbf{X}}^p_i)-MP(z_{\mathbf{\omega}}(\hat{\mathbf{X}}^p_i))||^2
\end{split}
\label{eq7}
\end{equation}
where $\mathbf{y}\in \mathbb{R}^{q_s^{tr}}$ is the ground truth bag-label vector of one task, $q_s^{tr} + q_s^{val} = q_s$,  $q_s^{tr}$ and $q_s^{val}$ are the label space sizes of support set and query set in a source task, respectively.    $g_{\mathbf{\omega}}(\hat{\mathbf{X}}^p_i) = MP(Soft(MLP\mathbf(\hat{\mathbf{X}}^p_i))$ predicts the labels of bags.  $MLP(\cdot)$ is  a three-layer Multi-Layer Perceptron, $Soft(\cdot)$ is the Softmax operation and $MP(\cdot)$ is the column-wise Max Pooling operation to aggregate the labels of instances to their hosting bags. $g_{\mathbf{\omega}}$ predicts the labels of bags and is parameterized by $\mathbf{\omega}$. $z_{\mathbf{\omega}}(\hat{\mathbf{X}}^p_i)$ predicts the labels of instances. So the first term on the right of Eq. \eqref{eq7} pursues the label prediction of bags by referring to known labels of bags. The second term pursues the consistent labels from the bag-level and instance-level, and distributes the bag-level labels to individual instances of $\hat{\mathbf{X}}^p_i$.

To obtain the prior information on  meta-path $p$ from the global prior $\phi$, we perform gradient descent on loss in the meta-path $p$ induced semantic space for task $\mathcal{T}_i^s$. Here $\phi_i^p=\{{\theta}_i^p,{\omega}_i^p\}$, and ${\theta}_i^p$ indicate semantic-specific parameters and  ${\omega}_i^p$ denotes task-specific parameters.
In this way, MetaMIML can learn multiple aspects of meta-knowledge as follows:
\begin{equation}
 \begin{split}
{\theta}_i^p &= \theta - {\alpha}\frac{\partial \mathcal{L}_{\mathcal{T}_i}(\mathbf{\omega},\hat{\mathbf{X}}^p_i,\mathcal{G}_{i}^{tr,dir})}{\partial\theta}\\
&= \theta - {\alpha}\frac{\partial \mathcal{L}_{\mathcal{T}_i}(\mathbf{\omega},\hat{\mathbf{X}}^p_i,\mathcal{G}_{i}^{tr,dir})}{\partial\hat{\mathbf{X}}^p_i} \frac{\partial \hat{\mathbf{X}}^p_i}{\partial\theta}
\end{split}
\label{eq8}
\end{equation}
where ${\alpha}$ is the learning rate. Different meta-path spaces encode different semantic information,  {and the semantic information extracted by different tasks are of different importance and relevance}. To gather different semantic contexts, we apply attention mechanism on different meta-paths to reduce the redundancy as follows:
\begin{equation}
{a}_i^p = Soft(\mathcal{L}_{\mathcal{T}_i}(\mathbf{\omega},\hat{\mathbf{X}}^p_i,\mathcal{G}_{i}^{tr,dir}))
\label{eq9}
\end{equation}
where ${a}_i^p$ is the weight of semantic space $p$ for task $\mathcal{T}_i^s$. This attention strategy also retains different semantics while quickly adapts to a new task. In this way, we can get the attention weights of all meta-paths $\mathcal{P}$.

To achieve cross-task knowledge across multiple meta-paths $\mathcal{P}$, we set ${\omega}_i = {\sum}_{p\in\mathcal{P}}{a}_i^p{\omega}_i^p$  and $\hat{\mathbf{X}}_i = {\sum}_{p\in\mathcal{P}}{a}_i^p\hat{\mathbf{X}}_i^p$.
Next, the task-wise can be adapted by the global prior knowledge ${\omega}$ and attention based ${\omega}_i$, which help predicting the labels of bags. The prediction task $\mathcal{T}_i$ with meta-path $p$  is adapted as follows:
\begin{equation}
{\omega}_i^p = {\omega}\otimes{\omega}_i - {\beta}\frac{\partial \mathcal{L}_{\mathcal{T}_i}(\mathbf{\omega}_i^p,\hat{\mathbf{X}}_i,\mathcal{G}_{i}^{tr,dir})}{\partial{\omega}_i^p}
\label{eq10}
\end{equation}
where ${\beta}$ is the learning rate and $\otimes$ is the element-wise product. As shown in Fig. \ref{fig1}(ii), the global prior $ \phi= \{\theta,\omega\}$ can then be optimized through query loss:
\begin{equation}
\underset{\phi}{min} \underset{\mathcal{T}_i^s\in \mathcal{D}^{s}}{\mathbb{E}} \mathcal{L}_{\mathcal{T}_i}(\mathbf{\omega}_i^p,\hat{\mathbf{X}}_i, \mathcal{G}_i^{val,dir})
\label{eq11}
\end{equation}
{Note the label space of a query bag  is $\mathbf{y} \in \mathbb{R}^{q_s^{val}}$,   $q_s^{val}$ is the number of distinct labels in a query set of a source task.}

To this end, MetaMIML not only learns different semantic representations of bags/instances via different meta-paths $p$, and generates multiple meta-training tasks from these semantic representations to acquire prior knowledge, but also can quickly adapt to new tasks from task-wise by prior knowledge using one (or a few) gradient descent step. The pseudo code of the training procedure  for MetaMIML is detailed in Algorithm \ref{alg:algorithm1}.

\begin{algorithm}[tb]
\caption{ Pseudo-code of MetaMIML }
\label{alg:algorithm1}
\textbf{Input}: a HMIN $\mathcal{G}$, a set of meta-path $\mathcal{P}$ and meta-training tasks $\mathcal{D}_{s}$; meta-learning rates: $\gamma$, $\alpha$ and $\beta$;\\
\textbf{Output}: meta knowledge $\phi=\{\theta,\omega\}$.
\begin{algorithmic}[1] 
\STATE Randomly initialize parameters $\phi = \{\theta,\omega\}$ and other global parameters (such as lookup table $\mathbf{W}$ and sparse projection matrix $\mathbf{E}$);
\STATE Construct the support set $\mathcal{G}_{i}^{tr}=  (\mathcal{G}_{i}^{tr,dir} , \mathcal{G}_{i}^{tr,\mathcal{P}})$ and query set $\mathcal{G}_{s}^{val} = (\mathcal{G}_{i}^{val,ind})$ for each task $\mathbf{X}_i$;
\WHILE{not converge}
\STATE Sample batch of bags $\mathcal{X}_k\sim \mathcal{D}_{s}$
\FOR{ $\mathbf{X}_i \in \mathcal{X}_k$}
\FOR{ ${p}\in \mathcal{P}$}
\STATE ${\theta}_i^p = {\theta}$, set ${\omega}_i^p = {\omega}$;
\STATE Generate $\hat{\mathbf{X}}_{i}^p$ by Eq. (5);
\FOR{ $\mathbf{x}_{i,j}$ in $\mathbf{X}_i$}
\STATE Generate $\hat{\mathbf{x}}_{i,j}^p$ by Eq. (6);
\ENDFOR
\STATE $\hat{\mathbf{X}}_i^p = \{\hat{\mathbf{x}}_{i,1}, \hat{\mathbf{x}}_{i,2},\cdots,\hat{\mathbf{x}}_{i,n_i}\}$;
\STATE Evaluate ${\nabla}_{\omega}\mathcal{L}_{\mathcal{T}_i}(\mathbf{\omega},\hat{\mathbf{X}}^p_i, \mathcal{G}_{i}^{tr,dir})$;
\STATE Locally update ${\theta}_i^p$ by Eq. (9);
\STATE Evaluate weight of semantics by Eq. (10);
\STATE Cross-task  knowledge  fusing ${\omega}_i = {\sum}_{p\in\mathcal{P}}{a}_i^p{\omega}_i^p$;
\STATE Update context fusion $\hat{\mathbf{X}}_i = {\sum}_{p\in\mathcal{P}}{a}_i^p\hat{\mathbf{X}}_i^p$;
\STATE Locally update ${\omega}_i^p$ by Eq. (11);
\ENDFOR
\ENDFOR
\STATE
 {$ \phi\leftarrow \phi- \gamma{\nabla}_{\phi}\underset{\mathcal{T}_i^s\in \mathcal{X}_k}{\mathbb{E}} \mathcal{L}_{\mathcal{T}_i}(\mathbf{\omega}_i^p,\hat{\mathbf{X}}_i, \mathcal{G}_i^{val,dir})$ }
\ENDWHILE
\STATE \textbf{return} $\phi=\{\theta,\omega\}$
\end{algorithmic}
\end{algorithm}

We conduct a complexity analysis of our training procedure, which include the context learner, random embedding operation and task learner.  The complexity of context learner and task learner are both $\mathcal{O}(|\mathcal{T}_s|\cdot|\mathcal{P}|\cdot d \cdot d_l\cdot n_m)$, and the complexity of random embedding operation is $\mathcal{O}((d+d_l)\cdot k\cdot n_m)$. Thus the time complexity of MetaMIML is $\mathcal{O}(e[|\mathcal{T}_s|\cdot|\mathcal{P}|\cdot d \cdot d_l + (d+d_l)\cdot k\cdot n_m])$, where $e$ is the number of epochs, $|\mathcal{T}_s|$ and $|\mathcal{P}|$ are the number of meta-training tasks and meta-paths, respectively. $d$ and $d_l$ are the dimension of instance and semantic context. $k$ and $n_m$ are the dimension of instance embeddings and the number of instances.   $|\mathcal{P}|$, $d$ , $ d_l$ and $k$ are usually small and random embedding operation can carry out$\sqrt{n}$-fold speedup, {so the complexity of MetaMIML is about linear with the number of tasks, w.r.t. $\mathcal{O}(|\mathcal{T}_s|\cdot log(n_m))$. }

\section{Experimental Results and Analysis}
\label{sec:exp}

\subsection{Experimental Setup}
\textbf{Datasets}: We use six benchmark datasets: Isoform, LncRNA, Birds, MSRC v2, Letter Carroll and Letter Frost.
Isoform and LncRNA are two biological datasets naturally linked with multi-types of molecules, they were used to  predict the  associations between isoforms/lncRNAs and functions/diseases. The last four are widely-used MIML datasets \cite{forrest2012rank,huang2019fast}. Table \ref{tab1} gives the
statistics of these datasets.

\noindent\textbf{Baselines}: To comparatively evaluate the performance of our  MetaMIML, we compare it against eight  methods of different categories, which include three MIML algorithms (MIMLfast \cite{huang2019fast}, MIMLSVM \cite{Zhou2012MIML} and MIMLNN \cite{zhou2008miml}) for MIML objects, two data fusion solutions based on matrix factorization (MFDF \cite{Zitnik2015DFMF} and SelDFMF \cite{wang2019SelDFMF}) and three network embedding based methods (Metapath2vec\cite{dong2017metapath2vec}, HANE\cite{wang2019heterogeneous} and M4L-JMF \cite{yang2020M4L}) for linked objects as follows:
\begin{itemize}
\item \textbf{MIMLSVM}~\cite{Zhou2012MIML}  uses a bag transformation strategy to convert bags into single instances and then degenerates MIML into a single-instance multi-label learning task. We set $ratio=0.2$ (parameter $k$ is set to be 20\% of the number of training bags), $svm.para=0.2$ (the value of `gamma') and $svm.type=Linear$ for Isoform dataset;  $ratio=0.3$, $svm.para=0.2$ and $svm.type=Poly$ for four MIML datasets;
     and $cost=1$ (the value of $C$) for all datasets.

\item  \textbf{MIMLNN}~\cite{zhou2008miml} utilizes  two-layer neural network structure to replace the MLSVM\cite{min2007mlsvm} used in MIMLSVM, then predicts the labels of bags.  We follow the configuration $ratio=0.4$ and $\lambda=0.5$ for Isoform dataset; $ratio=0.2$ and $\lambda=0.3$ for four MIML datasets;

\item  \textbf{MIMLfast}~\cite{huang2019fast}  first constructs a low-dimensional subspace shared by all labels, then trains label specific linear models to optimize approximated ranking loss via stochastic gradient descent. It can naturally detect key instance for each label by exploiting label relations in a shared space and discovering sub-concepts for complicated labels. We follow the configuration $d = 300$ (dimension of the shared space), $norm\_up =10$ (norm of each vector), $step\_size=0.005$ (step size of SGD), $\lambda=0.0001$ and $num\_sub=20$ (number of sub concepts) for Isoform; $d = 100$, $norm\_up =10$, $step\_size=0.003$, $\lambda=0.0005$ and $num\_sub=5$ for  four MIML datasets; $opts.norm=1$, $opts.average\_size=10$ and $opts.average\_begin=0$  for all datasets.

\item  \textbf{DFMF}~\cite{Zitnik2015DFMF}  collaboratively factorizes block matrices of a heterogeneous information network into low-rank matrices and then reconstructs the target relational matrix to predict the relations between bags and labels.   We follow the configuration $d =300 $ for Isoform dataset, while $d =20 $ for MIML datasets and $d =170 $ for LncRNA dataset.
\item  \textbf{SelDFMF}~\cite{wang2019SelDFMF}  can select and integrate inter-relational data sources by assigning different weights to them. We set $\alpha = 10^6$, $d =300 $ and $nTypes = 5$ for Isoform datase , while $\alpha = 10^5$, $d = 20$ and $nTypes = 3$ for four MIML datasets and $\alpha = 10^5$, $d = 160$ and $nTypes = 5$ for the LncRNA dataset.

\item  \textbf{M4L-JMF}~\cite{yang2020M4L}  firstly uses multiple data matrices to separately store the attributes and multiple inter(intra)-associations of objects, and then jointly factorizes these matrices into low-rank ones to explore the latent representation of each bag and its instances.
    We set $\alpha = 10^6$, $\beta = 10^6$, $d =240 $ and $nTypes = 5$ for  Isoform dataset; $\alpha = 10^5$,  $\beta = 10^6$, $d =20 $  and $nTypes = 3$ for  four MIML  datasets, and $\alpha = 10^7$, $\beta = 10^6$, $d = 160$ and $nTypes = 5$ for the LncRNA dataset.

\item  \textbf{Metapath2vec}~\cite{dong2017metapath2vec} is a classic heterogeneous information network representation learning method, which samples meta-path based random walks,  then leverages a heterogeneous skip-gram model to perform node embedding. We set the length of random walks, the number of walks and the size of windows to $l=40$, $w = 10$ and $size = 4$ respectively for  Isoform and LncRNA datasets. These parameter also was set for MetaMIML for walking.

\item  \textbf{HANE}~\cite{wang2019heterogeneous} uses an  attention mechanism on Graph Convolutional Networks (GCN) to encode the structure and attribute information of nodes to generate high-quality embedding. We set $d = 270$ for  Isoform dataset, $d = 150$ for  LncRNA dataset.
\end{itemize}
The input parameters of these comparison methods are specified (or optimized) according to the recommendations of the authors in their codes or papers.
All the experiments are performed on a server with following configurations: CentOS 7.3, 256GB RAM, Intel Exon E5-2678 v3 and NVIDIA Corporation GK110BGL [Tesla K40s]. We implement the proposed MetaMIML with deep learning library PyTorch. The Python and PyTorch versions are 3.6.5 and 1.3.1, respectively. We perform experiments on the six benchmark datasets to quantitatively study the performance of the proposed MetaMIML,  and  compare  it  against  eight  representative  and  related approaches as follows:

The first three compared methods are Multi-Instance Multi-Label (MIML) solutions, while the two methods(DFMF and SelDFMF) address the fusion of linked multi-types objects via matrix factorization, two network embedding methods (Metapath2vec and HANE) disregard the bag-instance associations. M4L-JMF also use heterogeneous multi-instance information network (HMIN). The input parameters of these comparison methods are specified (or optimized) according to the recommendations of the authors in their code or papers.
For the proposed MetaMIML, we adopt Adaptive Moment Estimation (Adam) to optimize our MetaMIML. For all datasets, we use a batch size 32 and set the meta-learning rate to 0.005 ($\gamma = 0.005$). We set both the context learner and task learner learning rate to 0.005 ($\alpha=\beta=0.005$) for Isoform and LncRNA dataset, while $\alpha=\beta=0.003$ four MIML  datasets. We perform one step gradient descent update in both context learner and task learner adaptations. We use the meta-path set $\{GDG, GG_oG, GMGD\}$ and $\{GDG, GLG, GLDG\}$ for Isofrom and LncRNA datasets (G: gene, D: disease, L: LncRNA, M: miRNA and $G_o$: Gene Ontology) respectively to MetaMIML and Metapath2vec. The maximum number of epochs are set as $e = 100$ and $e = 80$, $k = 240$ and $k = 100$  for Isofrom and LncRNA, respectively.  $e = 20$ and $k = 8$ for Birds,  we set $e = 15$ and $k = 10$ for other MIML datasets.


\noindent\textbf{Evaluation metrics}: To quantify the performance of MetaMIML, we adopted four canonical evaluation metrics, namely the average area under the precision-recall curve (\emph{AUPRC}), the average area under the receiver operating curve (\emph{AUROC}), the average  F1-score (\emph{AvgF1}) of all classes, and Hamming Loss (\emph{HL}). Unlike other metrics, the smaller the value of \emph{HL}, the better the performance is, so we report \emph{1-HL}.  For each bag (instance), we use the top $K$ labels with the largest probability as the relevant labels of the bag (instance). Here $K$ is the average number of labels per bag/instance. We report the average results (10 random partitions of each dataset) and standard deviations for each method.

\begin{table*}[bt]
   \centering

		\begin{tabular}{c |r r r r r r r r}\hline
	Dataset&          Instances&              Bags&        Labels&          AvgBL&     AvgBI&  NodeObject Types  & Link Type  \\ \hline
Letter Frost&            565  &                 144&             26 &     3.6 &     3.9 & ---     & ---    \\
Letter Carroll &         717  &                 166&             26 &     3.9 &     4.3 & ---     &---   \\
MSRC v2 &              1,758  &                591&             23 &     2.5 &     1.0  & ---    & --- \\
Birds &                 10,232  &               548&             13 &     2.1 &     18.7 & ---   & ---   \\
      \hline
    Isoform&         76,244&                  8,000&        6,428&            16.9&       6.5 &     5&  5\\
     LncRNA&        --- &                        240&        412&            --- &        --- &      6&      9\\
     \hline
		\end{tabular}
		
   \caption{Statistics of datasets used for the experiments. \emph{avgBL} is the average number of labels per bag and \emph{avgBI} is the average number of instances per bag. 
   }
   \label{tab1}
\end{table*}

We conducted three types of experiments to study the performance of MetaMIML. In the first experiments, we apply MetaMIML on Isoform  dataset to study two questions: how does MetaMIML perform compared with related methods? How does MetaMIML benefit from HMIN. In the second experiments, we compare MetaMIML against traditional MIML methods on four benchmark MIML datasets.
{In the third experiment, we explore the performance of MetaMIML and network embedding methods on LncRNA dataset with linked objects of multi-types.}

\subsection{ Results on Isoform}
\label{sec:isoforms}
In the first experiment, we use the Isoform dataset composed of 5 types of objects (miRNAs(495), genes(8,000), isoforms(76,244), Gene  Ontology(6,428),  Disease Ontology(8,450)) and 5 link types, the  more  detailed  information can be found in \cite{yu2020isoform}. Follow the experimental protocol (recommend items to users) in \cite{hoyeop2019melu}, we divided the samples and labels into two groups: \emph{source} and \emph{target} as the meta-training and meta-test data. We randomly divided the Gene Ontology labels (functional annotations of genes) into two sets (approximately 8:2), the  first 80\% labels serve as \emph{source} tasks and the remaining as the \emph{target} tasks. Then we randomly selected 80\% of the bags (genes) as the \emph{source} bags and the rest as \emph{target} bags. For each bag, five labels are randomly selected as the the query set and rest as the support set from \emph{source} labels during meta-training.  The same way to split \emph{target} labels for each task (bag) during meta-test stage.

Because MIML methods (MIMLSVM. MIMLNN and MIMLfast) cannot directly handle linked objects of multi-types, we first project other objects (except Gene Ontology labels) towards the genes to form MIML Isoform data, and then apply these MIML methods. For matrix factorization based data fusion (except M4L-JMF) and heterogeneous network embedding methods, we adopt the transformation used in \cite{yang2020M4L}, disregarding the bag-instance associations.

\begin{table*}[tb]
   \centering
		\begin{tabular}{l l l l l}\hline
 \textbf{Method}&          \textbf{ AvgF1}&                  \textbf{AUROC}&                \textbf{AUPRC}& \textbf{1-HL}\\
\hline
MIMLSVM&                    .097$\pm$.002$\bullet$&     .506$\pm$.001$\bullet$&    .068$\pm$.011$\bullet$&   .961$\pm$.006$\bullet$ \\
MIMLNN&                    .132$\pm$.001$\bullet$&     .558$\pm$.006$\bullet$&     .114$\pm$0.006$\bullet$&   .971$\pm$.003$\bullet$ \\
MIMLfast&                    .244$\pm$.006$\bullet$&     .823$\pm$.003$\bullet$&    .316$\pm$.005$\bullet$&   .889$\pm$.007$\bullet$ \\
    DFMF&                    .042$\pm$.003$\bullet$&     .851$\pm$.003$\bullet$&    .495$\pm$.006&            .974$\pm$.002 \\
 SelDFMF&                    .041$\pm$.002$\bullet$&     .846$\pm$.002$\bullet$&    .498$\pm$.002$\circ$ &    .979$\pm$.001 \\
 M4L-JMF&                    .051$\pm$.003$\bullet$&     .861$\pm$.004$\bullet$&    .527$\pm$.005$\circ$ &     .981$\pm$.002$\circ$ \\
Metapath2vec&                .176$\pm$.004$\bullet$&     .712$\pm$.008$\bullet$&    .391$\pm$.007$\bullet$&   .932$\pm$.004$\bullet$ \\
    HANE&                    .251$\pm$.002$\bullet$&     .827$\pm$.003$\bullet$&    .411$\pm$.003$\bullet$&   .967$\pm$.003$\bullet$ \\
MetaMIML&                    .286$\pm$.001&              .872$\pm$.002&             .491$\pm$.002&            .976$\pm$.003 \\
      \hline
		\end{tabular}
   \caption{ Results of compared methods on Isoform. $\bullet/\circ$ indicates whether MetaMIML is statistically (pairwise $t$-test at $95\%$ significance level) superior/inferior to the other method.}
   \label{tab2}
\end{table*}

Table \ref{tab2} reports results of on Isoform dataset. From these results, we have some important observations:\\
(i) MetaMIML has clear better (or comparable) performance results than most competitive methods. This facts the effectiveness of our MetaMIML for fast adapting to new tasks using few training samples. We find that the fusion of multi-types objects in HMIN improves the representation of bags/instances and contributes to a significantly increased AvgF1 and AUROC by at least 3\% and 1.3\%. MIMLSVM and MIMLNN often have the lowest performance in terms of AUROC and AUPRC, that is because they both degenerate an MIML problem into a single-instance multi-label learning problem, which causes information loss. None of these compared methods consistently holds the best performance across the four evaluation metrics, that is because these metrics quantify the performance of multi-label learning from different perspectives, and a learner can not always outperform another one across all the metrics. This fact also signifies the complexity and hardship of MIML problems.


\noindent (ii) Heterogeneous multi-instance information network helps the fusion of linked objects of multi-types. This is supported by classical MIML methods have a lower performance than the network embedding based methods. The former cannot utilize the network structure information. MIMLfast is relatively better than the other two  MIML methods (MIMLSVM and MIMLNN), since it trains label specific linear models to optimize approximated ranking loss by stochastic gradient descent. However, MIMLfast can not model linked objects of different types. So its often loses to data fusion based solutions. Metapath2vec also performs meta-path based random  walks  and uses heterogeneous skip-gram model to learn the representation of bags, but it disregards the instance information. As such, it beats by M4L-JMF and MetaMIL, which consider the bag-instance associations. For the similar reason, matrix factorization based data fusion methods (DFMF and SelDFMF) also often have a lower performance than MetaMIML.
 Although M4L-JMF can fuse multi-types of objects and model the bag-instance associations, its avgF1 and AUROC values are lower than MetaMIML, since it can not extract and use meta-knowledge for fast adapting to new tasks  M4L-JMF has higher values of AUPRC and 1-HL than MetaMIML, since it uses an aggregation term to  push  the  labels  of  bags  to  their  affiliated  instances in a coherent manner. HANE uses an  attention mechanism with graph convolutional networks to mine the structure and attribute information of network nodes, but it loses to MetaMIML by a large margin, due to its inability to extract meta-knowledge for new tasks.

In summary, MetaMIML can not only utilize HMIN to fuse multi-types of objects, but also can leverage context learner and task leaner to extract meta knowledge from multiple semantic contexts induced by different meta-paths, and fast adapt to new tasks. For these advantages, it frequently outperforms those competitive compared methods.
\begin{table*}[btp]
\centering
    \begin{tabular}{c|c c c c c c| c }
    \hline
    \textbf{Metric} &\textbf{MIMLfast} &\textbf{MIMLSVM} &\textbf{MIMLNN} &\textbf{M4L-JMF}  &\textbf{DFMF} &\textbf{SelDFMF} &\textbf{MetaMIML} \\
    \hline
     & \multicolumn{7}{c}{\textbf{\textit{Birds}}}\\
     \cline{2-8}

     AvgF1 &.295$\pm$.005$\bullet$
     &.525$\pm$.002$\bullet$
     &.616$\pm$.003$\circ$
     &.268$\pm$.014$\bullet$
     &.252$\pm$.012$\bullet$
     &.261$\pm$.009$\bullet$
     &.575$\pm$.003\\
     AUROC &.518$\pm$.002$\bullet$
     &.743$\pm$.002$\bullet$
     &.774$\pm$.005$\bullet$
     &.944$\pm$.002$\circ$
     &.902$\pm$.009
     &.912$\pm$.003$\circ$
     &.906$\pm$.002\\
     AUPRC&.235$\pm$.000$\bullet$
     &.535$\pm$.007$\bullet$
     &.713$\pm$.013$\bullet$
     &.963$\pm$.008
     &.886$\pm$.045$\bullet$
     &.891$\pm$.018$\bullet$
     &.965$\pm$.003\\
     1-HL  &.756$\pm$.005$\bullet$
     &.846$\pm$.007$\bullet$
     &.865$\pm$.004$\bullet$
     &.972$\pm$.001$\bullet$
     &.971$\pm$.001$\bullet$
     &.972$\pm$.000$\bullet$
     &.985$\pm$.002\\
     \hline
      & \multicolumn{7}{c}{\textbf{\textit{Letter Carroll}}}\\
     \cline{2-8}

     AvgF1 &.349$\pm$.004$\bullet$
     &.369$\pm$.003$\circ$
     &.374$\pm$.005$\circ$
     &.288$\pm$.014$\bullet$
     &.247$\pm$.014$\bullet$
     &.269$\pm$.027$\bullet$
     &.366$\pm$.002\\
     AUROC &.753$\pm$.001$\bullet$
     &.623$\pm$.007$\bullet$
     &.626$\pm$.005$\bullet$
     &.924$\pm$.014
     &.906$\pm$.008$\bullet$
     &.921$\pm$.004
     &.924$\pm$.002\\
     AUPRC&.329$\pm$.001$\bullet$
     &.469$\pm$.002$\bullet$
     &.518$\pm$.002$\bullet$
     &.948$\pm$.008
     &.909$\pm$.056$\bullet$
     &.913$\pm$.027$\bullet$
     &.948$\pm$.008\\
     1-HL  &.726$\pm$.005$\bullet$
     &.829$\pm$.001$\bullet$
     &.834$\pm$.003$\bullet$
     &.982$\pm$.001$\bullet$
     &.981$\pm$.001$\bullet$
     &.983$\pm$.001$\bullet$
     &.986$\pm$.001\\
     \hline
      & \multicolumn{7}{c}{\textbf{\textit{Letter Frost}}}\\
     \cline{2-8}

     AvgF1 &.282$\pm$.002$\bullet$
     &.354$\pm$.001$\bullet$
     &.401$\pm$.011$\bullet$
     &.250$\pm$.014$\bullet$
     &.242$\pm$.018$\bullet$
     &.246$\pm$.029$\bullet$
     &.452$\pm$.009\\
     AUROC &.749$\pm$.000$\bullet$
     &.583$\pm$.005$\bullet$
     &.576$\pm$.006$\bullet$
     &.924$\pm$.002$\bullet$
     &.907$\pm$.003$\bullet$
     &.912$\pm$.004$\bullet$
     &.934$\pm$.002\\
     AUPRC&.325$\pm$.000$\bullet$
     &.472$\pm$.007$\bullet$
     &.534$\pm$.016$\bullet$
     &.951$\pm$.002$\circ$
     &.894$\pm$.049
     &.895$\pm$.029
     &.926$\pm$.008\\
     1-HL  &.787$\pm$.001$\bullet$
     &.854$\pm$.012$\bullet$
     &.867$\pm$.005$\bullet$
     &.985$\pm$.001
     &.983$\pm$.001$\bullet$
     &.983$\pm$.001$\bullet$
     &.985$\pm$.001\\
     \hline
      & \multicolumn{7}{c}{\textbf{\textit{MSRC v2}}}\\
     \cline{2-8}

     AvgF1 &.249$\pm$.001$\bullet$
     &.548$\pm$.005$\circ$
     &.462$\pm$.002$\bullet$
     &.215$\pm$.004$\bullet$
     &.198$\pm$.008$\bullet$
     &.213$\pm$.007$\bullet$
     &.537$\pm$.003\\
     AUROC &.831$\pm$.000$\bullet$
     &.875$\pm$.008$\bullet$
     &.867$\pm$.006$\bullet$
     &.958$\pm$.001$\circ$
     &.937$\pm$.002$\bullet$
     &.939$\pm$.001$\bullet$
     &.942$\pm$.001\\
     AUPRC&.311$\pm$.001$\bullet$
     &.712$\pm$.008$\bullet$
     &.724$\pm$.002$\bullet$
     &.933$\pm$.003$\circ$
     &.880$\pm$.021$\bullet$
     &.905$\pm$.009$\bullet$
     &.913$\pm$.001\\
     1-HL &.912$\pm$.001$\bullet$
     &.916$\pm$.002$\bullet$
     &.925$\pm$.003$\bullet$
     &.982$\pm$.001$\bullet$
     &.981$\pm$.000$\bullet$
     &.981$\pm$.001$\bullet$
     &.986$\pm$.001\\
     \hline
  \end{tabular}
      \caption{Bag-level results of compared methods on four datasets with known bag/ instance-level labels.  $\bullet/\circ$ indicates whether MetaMIML is statistically (pairwise t-test at $95\%$ significance level) superior/inferior to the other method. }
\label{tab3}
\vspace{-0.2em}
\end{table*}

\subsection{Results on MIML data}
In the second experiment, we use four MIML datasets listed Table \ref{tab1} to study the performance of MetaMIML and related methods on typical MIML tasks. Here, we randomly partition the samples of each dataset into a training set (70\%) and a testing set (30\%). Following the protocol in \cite{xing2019M3Lcmf}, we construct a heterogeneous network of bags and instances, and then use this network as input for matrix factorization/network embedding based compared methods.
Table \ref{tab3} reports the bag-level evaluation results.  Compared with other methods, MetaMIML often has the best performance on each dataset, and improves the AUPRC and AUPRC more obviously.

We report the experimental results  at instance-level in Table \ref{tabS1}, MetaMIML has the highest AvgF1 values, and MetaMIML improves AvgF1 by at least 14\% compared with MIMLfast. That is because MetaMIML can learn a good representation for instances.
M4L-JMF has a higher AUROC values than MetaMIML on Birds and MSRC v2 datasets. That is because M4L-JMF not only can push  the  labels  of  bags  to  their  affiliated  instances, but also can reversely aggregate the labels of instances to their hosting bags. But compared with the results at bag-level and at nstance-level, MetaMIML  beats M4L-JMF in terms of other evaluation metrics. This fact proves that MetaMIML can also make a competitive instance-level label prediction. The instance-level evaluation results  also prove that MetaMIML outperforms compared methods, due to the improved instance representation via fusion of bag-level context semantics.

\begin{table}[h!tbp]
\centering
    \begin{tabular}{c|l  l   | l }
    \hline
    \textbf{Metric}   &\textbf{MIMLfast}   &\textbf{M4L-JMF} &\textbf{MetaMIML} \\
    \hline
     & \multicolumn{3}{c}{\textbf{\textit{Birds}}}\\
     \cline{2-4}

     AvgF1 &0.228$\pm$0.001$\bullet$
     &0.291$\pm$0.017$\bullet$
     &0.372$\pm$0.014
     \\
     AUROC &0.512$\pm$0.002$\bullet$
     &0.957$\pm$0.018$\circ$
     &0.891$\pm$0.006
     \\
     AUPRC &0.395$\pm$0.008$\bullet$
     &0.961$\pm$0.004$\bullet$
     &0.978$\pm$0.003
     \\
     1-HL &0.726$\pm$0.014$\bullet$
     &0.891$\pm$0.006$\bullet$
     &0.924$\pm$0.002
     \\
     \hline

      & \multicolumn{3}{c}{\textbf{\textit{Letter Carroll}}}\\
     \cline{2-4}

     AvgF1 &0.113$\pm$0.009$\bullet$
     &0.085$\pm$0.005$\bullet$
     &0.253$\pm$0.003
     \\
     AUROC &0.517$\pm$0.012$\bullet$
     &0.891$\pm$0.031
     &0.887$\pm$0.013
     \\
     AUPRC &0.308$\pm$0.009$\bullet$
     &0.913$\pm$0.014
     &0.912$\pm$0.002
     \\
     1-HL &0.715$\pm$0.007$\bullet$
     &0.914$\pm$0.003$\bullet$
     &0.936$\pm$0.005
     \\
     \hline

      & \multicolumn{3}{c}{\textbf{\textit{Letter Frost}}}\\
     \cline{2-4}

     AvgF1 &0.262$\pm$0.014$\bullet$
           &0.072$\pm$0.005$\bullet$
           &0.411$\pm$0.004
     \\
     AUROC &0.512$\pm$0.002$\bullet$
           &0.894$\pm$0.011
           &0.886$\pm$0.005
     \\
     AUPRC &0.313$\pm$0.007$\bullet$
     &0.884$\pm$0.009$\bullet$
     &0.915$\pm$0.012
     \\
     1-HL &0.715$\pm$0.016$\bullet$
     &0.881$\pm$0.002$\bullet$
     &0.921$\pm$0.007
     \\
     \hline

           & \multicolumn{3}{c}{\textbf{\textit{MSRC v2}}}\\
     \cline{2-4}

     AvgF1 &0.195$\pm$0.011$\bullet$
            &0.147$\pm$0.002$\bullet$
            &0.318$\pm$0.003
     \\
     AUROC &0.541$\pm$0.006$\bullet$
           &0.885$\pm$0.013$\circ$
           &0.864$\pm$0.009
     \\
     AUPRC &0.271$\pm$0.004$\bullet$
     &0.867$\pm$0.021$\bullet$
      &0.889$\pm$0.012
     \\
     1-HL &0.782$\pm$0.007$\bullet$
     &0.912$\pm$0.005$\bullet$
     &0.934$\pm$0.006
     \\
     \hline
  \end{tabular}

      \caption{Results of MetaMIML and compared methods on four datasets with instance-level labels.  $\bullet/\circ$ indicates whether MetaMIML is statistically (according to pairwise t-test at $95\%$ significance level) superior/inferior to the other method. }

\label{tabS1}

\end{table}

In the third experiment, we further study the performance of MetaMIML on linked objects of different types. We use the LncRNA dataset \cite{wang2019SelDFMF}, which is composed of 6 types of objects (LncRNA(240), genes(15,527), miRNAs(495),  Gene  Ontology(6,428),  Disease Ontology(412) and Drug(8,283)) and 9 types of links.
We report  the results in  Table \ref{tabS2}. Overall, MetaMIML has a relatively stable performance and frequently outperforms competitive compared methods. These experiment results  not  only  demonstrate the flexibility of MetaMIML in diverse settings, but also prove the effectiveness of the context leaner for exacting the network structure information via different meta-paths.

\begin{table*}[h!tbp]
   \centering

\begin{tabular}{c l l l l}\hline
	\textbf{Method}&      \textbf{AvgF1}&              \textbf{AUROC}&                \textbf{AUPRC}&             \textbf{1-HL}\\
\hline
DFMF&                    0.062$\pm$0.001$\bullet$&      0.872$\pm$0.007$\bullet$&      0.546$\pm$0.091$\bullet$&   0.982$\pm$0.001  \\
SelMFDF&                 0.066$\pm$0.003$\bullet$&      0.887$\pm$0.003&               0.604$\pm$0.015&            0.986$\pm$0.001 \\
M4L-JMF&                 0.067$\pm$0.002$\bullet$&      0.895$\pm$.004$\circ$&        0.616$\pm$0.026&            0.989$\pm$.001$\circ$ \\
metapath2vec&            0.096$\pm$0.002$\bullet$&      0.771$\pm$0.013$\bullet$&      0.375$\pm$0.014$\bullet$&   0.952$\pm$0.006$\bullet$    \\
HANE&                    0.224$\pm$0.002$\bullet$&      0.867$\pm$0.008$\bullet$&        0.471$\pm$.008$\bullet$&   0.975$\pm$0.005$\bullet$    \\
MetaMIML&                0.282$\pm$0.003&               0.886$\pm$.002&               0.611$\pm$0.008&            0.984$\pm$.002$\bullet$  \\
      \hline
		\end{tabular}
   \caption{ Results of MetaMIML,  matrix factorization based and network embedding methods on the LncRNA dataset. $\bullet/\circ$ indicates whether MetaMIML is statistically (according to pairwise t-test at $95\%$ significance level) superior/inferior to the other method.  }
      \label{tabS2}
\end{table*}

Overall, these  results  also  confirm  that MetaMIML can also work on traditional MIML tasks.

\subsection{ Parameter Sensitivity Analysis}

\begin{figure}[tp]
\centering
\subfigure[LncRNA ]{\includegraphics[height=4cm,width=4cm]{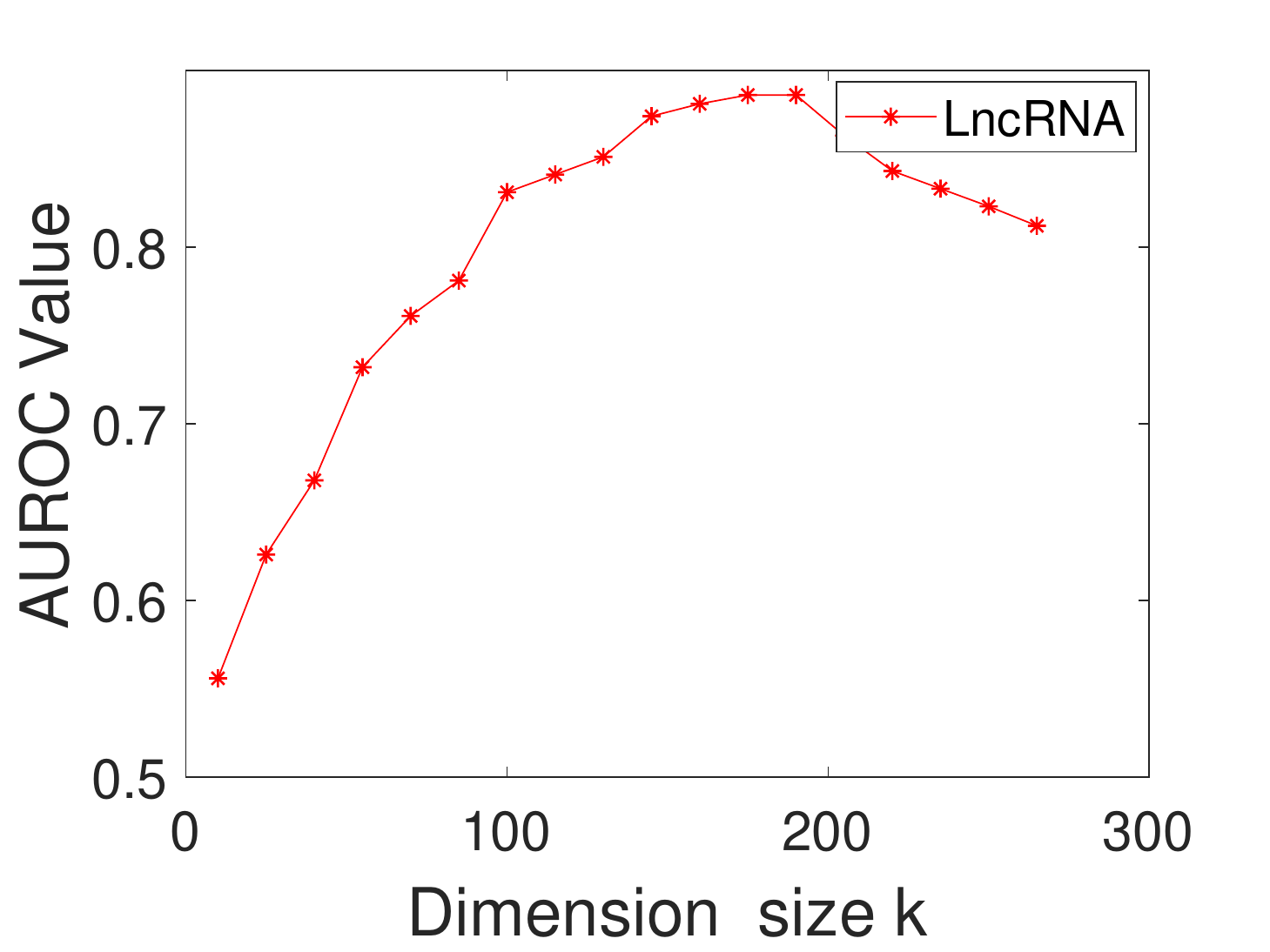}}\label{fig1a}
\subfigure[Letter Carroll]{\includegraphics[height=4cm,width=4cm]{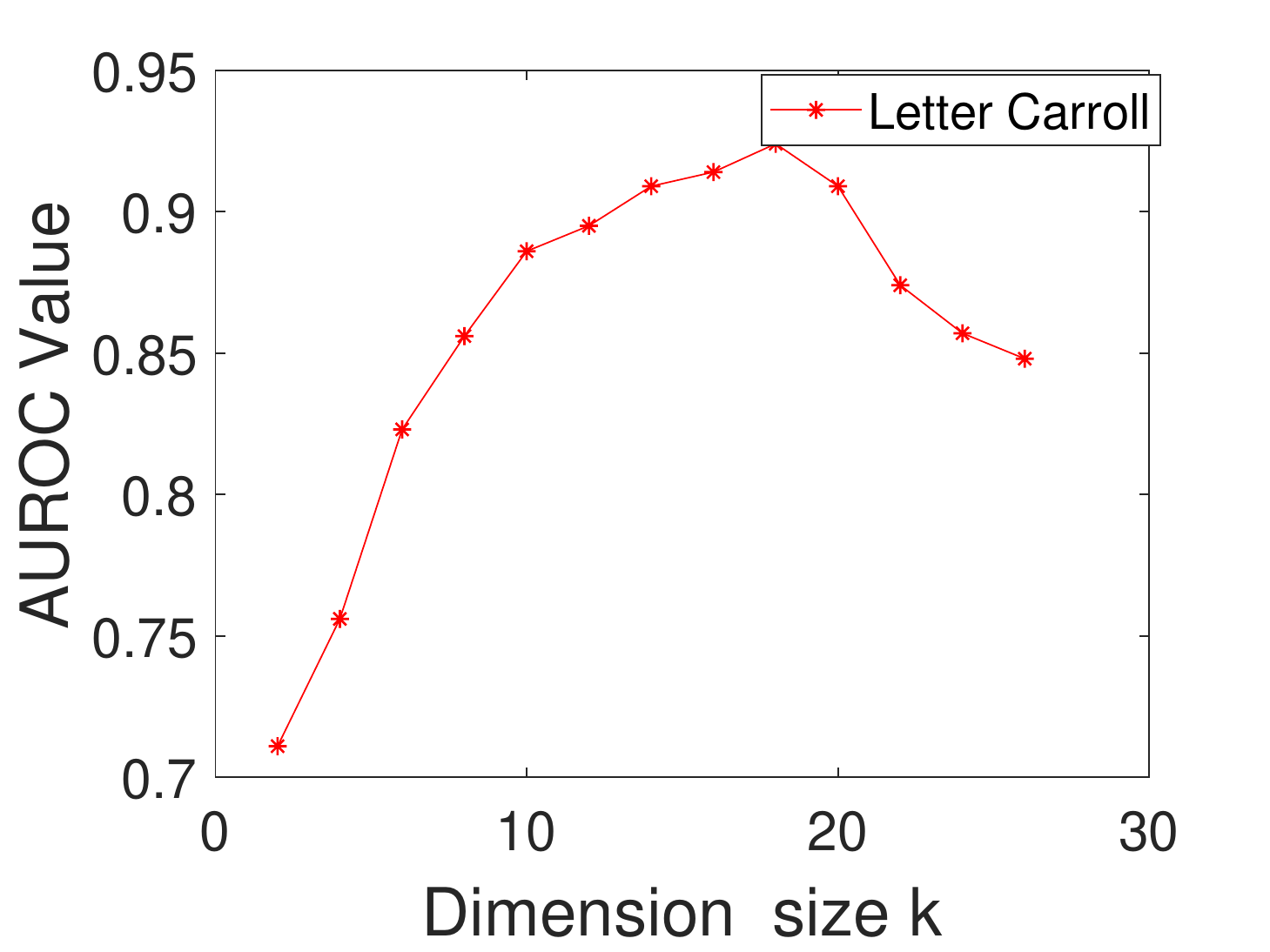}}\label{fig1b}
\caption{ AUROC vs. $k$ (dimension size of $\textbf{E}$) on the LncRNA and  Letter Carrol datasets.}
\label{fig1}
\end{figure}
In this paper,  parameter $k$ (the column size of sparse random projection embedding matrix $\mathbf{E}$) (in Eq . (6) of the main text) should be specified for our proposed MetaMIML.
We observe from Fig. \ref{fig1} that a too small $k$ can not sufficiently encode  multi-types objects, and while a too large $k$ brings in some noises and thus leads to a low AUROC value. From the above analysis, we adopt $k=155$ and $k=18$ for experiments on LncRNA and Letter Carroll, respectively.

\section{Conclusions}
\label{sec:concl}
In this paper, we investigate how to perform meta learning on multi-instance multi-label data, which is a challenging and practical,  but under-studied problem. We introduce an approach called  MetaMIML for naturally linked MIML objects of multi-types. Experimental results on real-world datasets show that MetaMIML can quickly adapt to new tasks and achieve a better performance than other competitive and related methods. MetaMIML also outperforms these methods in the typical MIML tasks.

\bibliographystyle{named}
\bibliography{ijcai21}

\end{document}